\documentclass[%
 aip,
 amsmath,amssymb,
reprint,%
]{revtex4-1}

\usepackage{graphicx}
\usepackage{subcaption} 
\usepackage{dcolumn}
\usepackage{bm}

\usepackage[usenames,dvipsnames]{xcolor}
\usepackage[utf8]{inputenc}
\usepackage[T1]{fontenc}
\usepackage{mathptmx}
\usepackage{etoolbox}
\usepackage[hidelinks]{hyperref}
\usepackage{array}
\usepackage{multirow}
\usepackage{amsmath, amssymb, amsthm, amsbsy}
\usepackage{amstext}
\usepackage{natbib}
\usepackage{graphicx, caption}
\usepackage{subcaption}
\usepackage{hyperref}  

\makeatletter
\def\@email#1#2{%
 \endgroup
 \patchcmd{\titleblock@produce}
  {\frontmatter@RRAPformat}
  {\frontmatter@RRAPformat{\produce@RRAP{*#1\href{mailto:#2}{#2}}}\frontmatter@RRAPformat}
  {}{}
}%
\makeatother

\usepackage[
lined,
ruled, 
vlined,
commentsnumbered, 
]{algorithm2e}

\usepackage{svg}

\newcommand{\jacobian}[1]{\mathbf{J}\left(#1\right)}
\newcommand{\R}{\mathbb{R}}
\newcommand{\dx}{\text{d}x}
\newcommand{\dy}{\text{d}y}
\newcommand{\dr}{\mathrm{d}}
\newcommand{\mbf}{\mathbf}
\newcommand{\eqsplit}[1]{\begin{equation}\begin{split}#1\end{split}\end{equation}}
\newcommand{\eqsplits}[1]{\begin{align*}#1\end{align*}}

\newcommand{\E}{\mathbb{E}}

\newtheorem{theorem}{Theorem}
\newtheorem{definition}{Definition}[section]

\newcommand{\duoftcs}{Department of Computer Science, University of Toronto, Toronto, Ontario M5S 2E4, Canada}
\newcommand{\dvector}{Vector Institute for Artificial Intelligence, Toronto, Ontario M5G 1M1, Canada}
\newcommand{\duoftc}{Department of Chemistry, University of Toronto, Toronto, Ontario M5S 3H6, Canada}
\newcommand{\dacc}{Acceleration Consortium, Toronto, Ontario M5S 3H6, Canada}
\newcommand{\duoftac}{Department of Chemical Engineering \& Applied Chemistry, University of Toronto, Toronto, Ontario M5S 3E5, Canada}
\newcommand{\duoftmse}{Department of Materials Science \& Engineering, University of Toronto, Toronto, Ontario M5S 3E4, Canada}
\newcommand{\dcifar}{Lebovic Fellow, Canadian Institute for Advanced Research (CIFAR), Toronto, Ontario M5G 1M1, Canada}

\begin{document}
\preprint{AIP/123-QED}

\title{Waveflow: boundary-conditioned normalizing flows applied to fermionic wavefunctions}

\author{Luca Thiede}
\thanks{These authors contributed equally.}
\affiliation{\duoftcs}
\affiliation{\dvector}
\author{Chong Sun}
\thanks{These authors contributed equally.}
\affiliation{\duoftcs}
\email{chongs0419@gmail.com}

\author{Al\'{a}n Aspuru-Guzik}
\affiliation{\duoftcs}
\affiliation{\dvector}
\affiliation{\duoftc}
\affiliation{\duoftac}
\affiliation{\duoftmse}
\affiliation{\dacc}
\affiliation{\dcifar}
\email{alan@aspuru.com}

\date{\today}

\begin{abstract}
An efficient and expressive wavefunction ansatz is key to scalable solutions for complex many-body electronic structures.
While Slater determinants are predominantly used for constructing antisymmetric electronic wavefunction ans\"{a}tze, this construction can result in limited expressiveness when the targeted wavefunction is highly complex.
In this work, we introduce Waveflow, an innovative framework for learning many-body fermionic wavefunctions using boundary-conditioned normalizing flows. 
Instead of relying on Slater determinants, Waveflow imposes antisymmetry by defining the fundamental domain of the wavefunction and applying necessary boundary conditions. 
A key challenge in using normalizing flows for this purpose is addressing the topological mismatch between the prior and target distributions. 
We propose using O-spline priors and I-spline bijections to handle this mismatch, which allows for flexibility in the node number of the distribution while automatically maintaining its square-normalization property.
We apply Waveflow to a one-dimensional many-electron system, where we variationally minimize the system's energy using variational quantum Monte Carlo (VQMC).
Our experiments demonstrate that Waveflow can effectively resolve topological mismatches and faithfully learn the ground-state wavefunction. 
\end{abstract}
\maketitle

\section{Introduction\label{sec:intro}}
Scalable solutions to quantum chemistry are actively pursued through numerical modeling\cite{szabo1982modern, levine2014quantum, jensen2017introduction}, quantum computing\cite{cao2019quantum, motta2021emerging}, and machine learning\cite{goh2017deep, dral2020quantum, hermann2023ab}. 
While the electronic structures in chemical systems are governed by the many-body Schr\"{o}dinger equation, exact solutions to it are feasible only for small systems due to the exponential growth of the computational complexity with the number of electrons. 
Traditional quantum chemistry methods use an \textit{ab initio} approach, making no approximations other than the Born-Oppenheimer approximation to the Hamiltonian, which separates the electronic and nuclear motion.
Numerical approximations are made by selecting a function that describes the electronic structure, known as the wavefunction. 
A proposed form of wavefunctions is called an \textit{ansatz}, which usually comes with a set of adjustable parameters.
Throughout this paper, we use the term ans\"{a}tze (the plural form of ansatz) to refer to multiple trial functions or models.
Well-known ans\"{a}tze include the Hartree-Fock (HF), configuration interaction (CI)~\cite{Foresman1992CIS, Headgordon1994CISD}, coupled cluster (CC)~\cite{Cizek1966CC, Purvis1982CCSD}, and matrix product state (MPS)~\cite{White1992Density, Orus2014AnnalPhys, orus2019tensor} ans\"{a}tze. 
Different numerical algorithms are then developed based on the choice of the ansatz.
For example, using the HF ansatz reduces the two-electron interaction into an effective one-electron potential, which can be solved by the self-consistent field (SCF) method. 
The MPS ansatz offers a quantitative description of the entanglement in one-dimensional systems and can be systematically simplified by iteratively truncating the less-entangled parts using methods such as the density matrix renormalization group (DMRG)\cite{White1992Density}. 
However, these traditional quantum chemistry methods face significant accuracy-efficiency tradeoffs.
Additionally, they rely heavily on physical intuition, which can limit the expressiveness of the wavefunctions and the ability to achieve accurate results for complex systems.

\begin{figure*}[t!]
    \centering
    \includegraphics[width=0.9\textwidth]{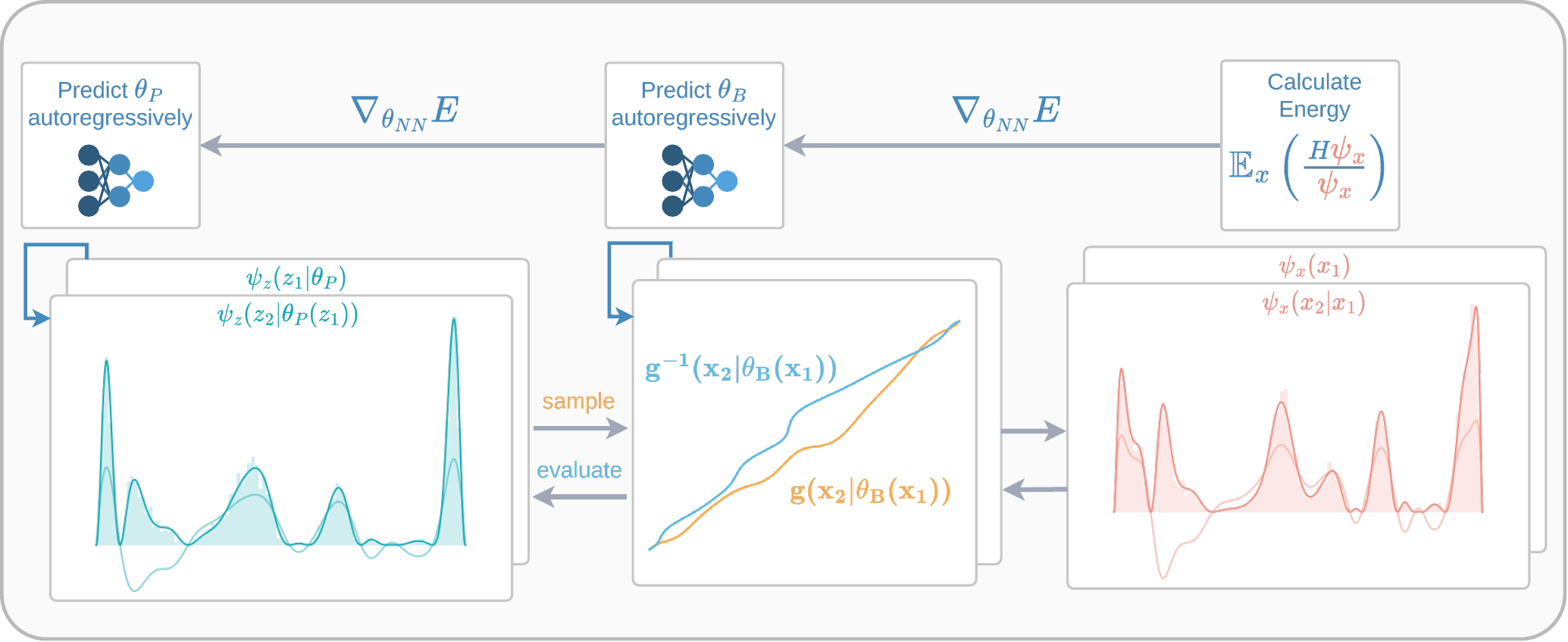}
    \caption{Learning the ground-state fermionic wavefunction using Waveflow and variational quantum Monte Carlo (VQMC).}
    \label{fig:toc}
\end{figure*}

The recent development of neural network quantum states (NNQS) has opened new horizons for designing many-body wavefunction ans\"{a}tze and corresponding algorithms~\cite{hermann2023ab, medvidovic2024neural}.
Unlike the traditional quantum chemistry ans\"{a}tze, neural networks possess high expressiveness and thus can potentially capture complex correlations in many-body systems.
Additionally, the structures of neural networks provide efficient high-dimensional optimization and massive parallelization using GPUs.  
Initial NNQS trials focused on spin systems~\cite{Carleo2017Solving, Sharir2020Deep, Hibat2020Recurrent}. 
However, expanding NNQS from spin systems to electronic systems is challenging. 
Unlike spins, which commute with each other, electrons are spin-$\frac{1}{2}$ fermions governed by Fermi-Dirac statistics and thus anticommute.
This requires the electronic wavefunction to gain a minus sign upon the exchange of two electrons, resulting in electronic wavefunctions being antisymmetric.
One approach to treating an electronic Hamiltonian involves transforming it into a spin Hamiltonian using fermion-to-spin transformations such as the Jordan-Wigner (JW) transformation~\cite{choo2020fermionic} and the Bravyi-Kitaev transformation~\cite{bravyi2002fermionic}. 
However, these transformations typically introduce higher-order and long-range interactions.
Another approach is to use Slater determinants as the building blocks of electronic wavefunctions. 
A Slater determinant is an uncorrelated many-body wavefunction that satisfies antisymmetry. 
A correlated wavefunction can be written as a linear combination of many Slater determinants and maintains antisymmetry. 
One can enhance the flexibility of Slater determinants to some extent by correlating the single-particle orbitals using methods such as the backflow neural networks\cite{Ruggeri2018Nonlinear, luo2019backflow, hermann2020deep} and the hidden-fermion determinant states\cite{Moreno2022Fermionic}.
Nevertheless, requiring Slater determinants to be the building blocks can sometimes restrict the choice of ans\"{a}tze and thus limit its expressiveness. For example, most quantum chemistry methods start from a reference Slater determinant and add low-excitation determinants. 
If the reference determinant is not optimal, this approach can lead to poor accuracy or slow convergence, known as the single-reference problem.
In this work, we offer a different perspective by bypassing the requirement of using Slater determinants. 
Instead, we directly encode antisymmetry into a boundary-conditioned normalizing flow ansatz, aiming to provide a more flexible alternative.

An electronic wavefunction can be conceptualized as a square-normalized probability distribution. 
In the context of non-relativistic Hamiltonians, the wavefunction can always be chosen to be real. 
A natural candidate for generating complicated distributions are normalizing flows\cite{dinh2015nice, dinh2017density, papamakarios2017masked, kingma2018glow, chen2018neural,  durkan2019neural, papamakarios2021normalizing, Kobyzev2021normalizing}, which transform a simple prior distribution to a more complex distribution through diffeomorphic coordinate transformations. 
Normalizing flows ensure efficient sampling and exact likelihood estimation of complicated distributions. 
In addition, (square) normalization is automatically fulfilled by definition, which simplifies implementation and avoids numerical instabilities caused by diminishing or exploding wavefunction norms.
While most applications of normalizing flows focused on non-negative distributions, one can easily adapt these methods to handle square-normalized distributions by computing the square root of the Jacobian determinant~\cite{Xie2022ab}.
Previous applications of normalizing flows to physical and chemical systems focused on permutation-equivariant transformations\cite{noe2019boltzmann, kohler2020equivariant, kohler2021smooth}, while direct applications to many-electron wavefunctions remain largely unexplored. 
One challenge in employing normalizing flows for electronic wavefunctions is managing the number of nodes that separate the positive and negative amplitudes of the wavefunction. 
Due to the intricate nature of electronic structures, it is impractical to determine the number of nodes \textit{a priori}. 
On the other hand, if one starts from a prior distribution with a different number of nodes compared to the ground-truth distribution, normalizing flows fail to bridge the two distributions due to topological mismatches.

\begin{figure*}
    \centering
    \centering
    \begin{subfigure}[t]{0.3\textwidth}
      \centering
      \includegraphics[width=1\linewidth]{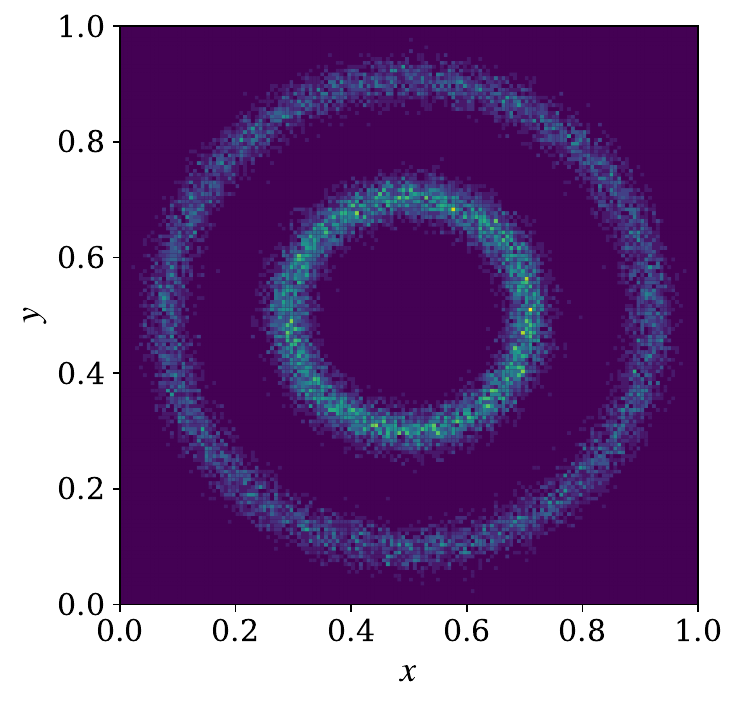}
      \caption{Target distribution}\label{fig:topological_two_circles_a}
    \end{subfigure}
    \begin{subfigure}[t]{0.3\textwidth}
      \centering
      \includegraphics[width=1\linewidth]{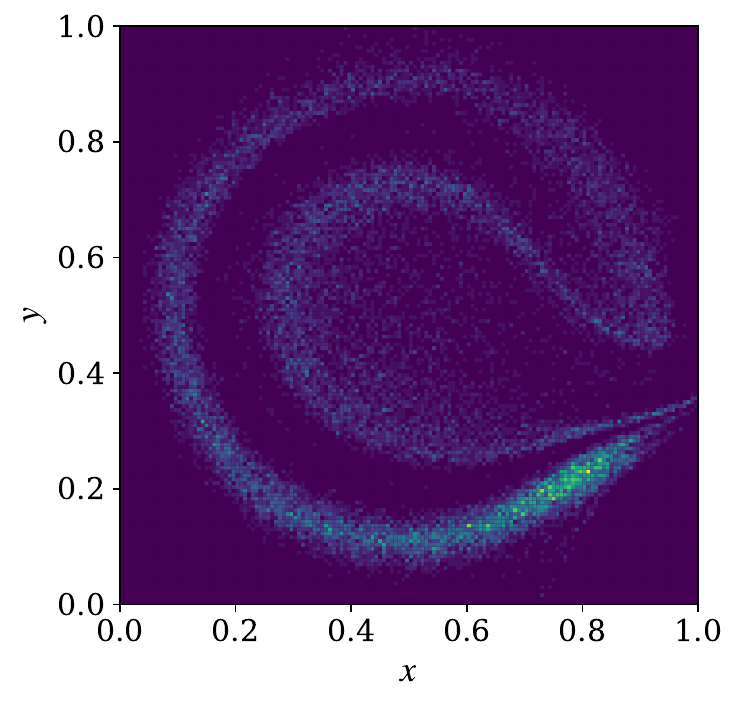}
      \caption{Affine coupling flow}
      \label{fig:topological_two_circles_b}
    \end{subfigure}
     \begin{subfigure}[t]{0.3\textwidth}
      \centering
      \includegraphics[width=1\linewidth]{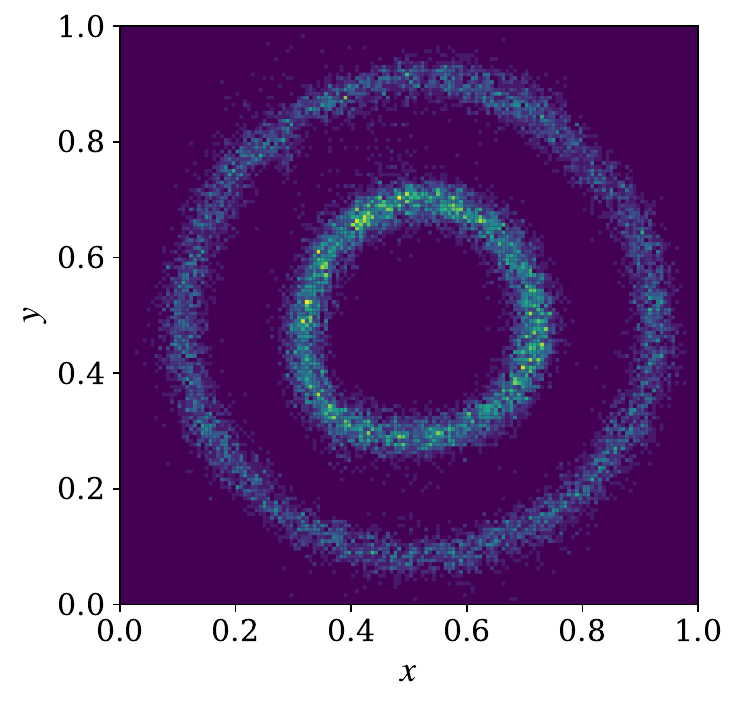}
      \caption{Waveflow}
      \label{fig:topological_two_circles_c}
    \end{subfigure}
    \caption{Topological mismatch with reproducing the double-circle distribution. (a) The target double-circle distribution.
    (b) Distribution learned by a normalizing flow using affine coupling layers.
    (c) Distribution learned by Waveflow.
    }
    \label{fig:topological_two_circles}
\end{figure*}

In this work, we present \textit{Waveflow}, a boundary-conditioned normalizing flow that can learn a complicated antisymmetric wavefunction in real space.
To encode antisymmetry, we define the fundamental domain of the permutation operators. 
We learn the wavefunction within this fundamental domain and enforce necessary boundary conditions~\cite{klimyk2007antisymmetric}.
These boundary conditions ensure compliance with the Pauli exclusion principle and maintain the continuity of the wavefunction throughout the entire space.
To address the topological mismatch problem, we introduce spline priors that are updated throughout the training process.
In addition, we use a spline flow to transform the distribution.
This allows us to make the bijection finite but arbitrarily differentiable, as well as to enforce the boundary conditions. 
We use variational quantum Monte Carlo (VQMC) to evaluate the system energy and energy gradient to train the Waveflow ansatz.
In VQMC, the sampling is usually performed with Markov chain Monte Carlo (MCMC), which can lead to biased sampling.
In contrast, normalizing flows enable efficient exact sampling from the target distribution, removing any bias.
This makes the Waveflow ansatz a natural fit for VQMC.
The methodology is detailed in Section~\ref{sec:method-general}.

We demonstrate the performance of Waveflow by reproducing disconnected distributions and evaluating the ground state of a one-dimensional helium-like system. 
One-dimensional electronic systems have served as the early test beds for various well-known electronic structure methods such as DMRG\cite{White1992Density} and quantum Monte Carlo (QMC)\cite{hirsch1982monte}. 
They are also actively studied in laboratories, including thin wires~\cite{roukes1987quenching}, carbon nanotubes\cite{peng2014carbon}, and edge states of two-dimensional topological materials\cite{halperin1982quantized, wada2011localized}.
One-dimensional systems often exhibit intrinsically different properties compared to higher-dimensional systems and have the potential for unique industrial applications~\cite{haldane2981effective, imambekov2012one,li2024imaging}.
In the experiments, we show that starting from a prior with an arbitrary number of nodes, Waveflow can correctly identify the topology of the ground state wavefunction. 
Therefore, our work not only addresses the topological issues associated with normalizing flows but also provides a framework for faithfully learning complex electronic wavefunctions. 
The details of the experiments are provided in Section~\ref{sec:experiments}.

\section{Methods\label{sec:method-general}}

In this section, we outline the key components for constructing Waveflow.
We begin with introductions to many-electron wavefunctions and ground-state energy estimation using variational quantum Monte Carlo (VQMC) in Sections~\ref{sec:electronicWavefunction} and \ref{sec:vqmc}.
Next, in Section~\ref{sec:method_normalizing_flows}, we summarize normalizing and square-normalizing flows, with detailed discussions on the choices of priors and bijections. 
We then address the topological mismatch problem and present our solution using the spline priors and spline bijections in Section~\ref{sec:topological_problem}.
Following this, we describe how to enforce boundary conditions and antisymmetry in normalizing flows in Section~\ref{sec:boundary_conditions}. 
Finally, we construct the Waveflow algorithm for one-dimensional many-electron systems in Section~\ref{sec:waveflow_alg}.
Fig.~\ref{fig:toc} provides an overview of our approach depicted visually.
In this work, we use the term "Waveflow" to represent our construction for both normalizing flows and square-normalizing flows, unless specified differently.

\subsection{Electronic wavefunction~\label{sec:electronicWavefunction}}

A many-electron wavefunction $\Psi(\mbf{r}_0, \mbf{r}_1, \ldots, \mbf{r}_{n-1})$ is a complex function that describes the quantum state of $n$ electrons in a system.
The square of its norm, $|\Psi(\mbf{r}_0, \mbf{r}_1, \ldots, \mbf{r}_{n-1})|^2$, represents the probability density of finding electron-0 at $\mbf{r}_0$, electron-1 at $\mbf{r}_1$, and so on. 
The wavefunction must satisfy the normalization condition:
\eqsplit{\label{eq:normalization_wavefunc}
\int  |\Psi(\mbf{r}_0, \mbf{r}_1, \ldots, \mbf{r}_{n-1})|^2 \dr \mbf{r}_0 \dr \mbf{r}_1 \ldots \dr \mbf{r}_{n-1} = 1,
}
which ensures that $\Psi(\mbf{r}_0, \mbf{r}_1, \ldots, \mbf{r}_{n-1})$ is {square-normalized}. 
In quantum chemistry, square normalization and normalization are often used interchangeably to indicate Eq.~\eqref{eq:normalization_wavefunc}. Here, we differentiate between the two definitions, as we will discuss both normalizing flows and square-normalizing flows in subsequent sections.

Electrons are spin-$\frac{1}{2}$ fermions~\cite{MannParticlePhys2010, Schwartz2013QFT}, meaning each electron has two spin freedoms: spin-up and spin-down. 
Therefore, $\mbf{r}_i$ should include both the spatial and spin coordinates. 
However, since we can decouple the wavefunctions into different spin components, in this work, we will focus only on the spatial coordinates $\mathbf{x}_i$ and assume all electrons are of the same spin.
We use $\psi(\mbf{x}_0, \mbf{x}_1, \ldots, \mbf{x}_{n-1})$ to represent a wavefunction that depends only on the spatial coordinates, often referred to as a spinless fermionic wavefunction. 
This wavefunction is antisymmetric with respect to exchanging the positions of two electrons:
\eqsplit{\label{eq:antisymm}
\psi( \ldots, \mbf{x}_i,\ldots, \mbf{x}_j, \ldots ) 
= -\psi( \ldots, \mbf{x}_j, \ldots, \mbf{x}_i, \ldots), \ i\neq j.
}
A consequence of antisymmetry is the Pauli exclusion principle, which prohibits two electrons from occupying the same quantum state. This results in the wavefunction $\psi$ being zero when two electrons coincide.
For simplicity, we use $\mathbf{x}$ to represent the collective spatial positions of all electrons in the following.

An ansatz is a specific form of functions defined by a set of parameters  $\bm{\theta}$, and is represented by $\psi_{\bm{\theta}}(\mathbf{x})$. 
The electronic ground state determines the static properties of a system, and it can be approximated by varying $\bm{\theta}$ to minimize the ground-state energy. In the next subsection, we present formulations for variationally approaching the best ground-state approximation given an ansatz $\psi_{\bm{\theta}}(\mathbf{x})$.

\subsection{Energy and gradient evaluation \label{sec:vqmc}} 

The energy of a quantum system described by $\psi_{\bm{\theta}}(\mathbf{x})$ is evaluated as
\eqsplit{\label{eq:expectation_hamiltonian}
E(\bm{\theta}) &= \int \psi^*_{\bm{\theta}}(\mathbf{x}) H 
\psi_{\bm{\theta}}(\mathbf{x})\ \dr \mathbf{x}\\
 &= \int |\psi_{\bm{\theta}}(\mathbf{x})|^2 \frac{H 
\psi_{\bm{\theta}}(\mathbf{x})}{\psi_{\bm{\theta}}(\mathbf{x})} \dr \mathbf{x}
= \int \rho_{\bm{\theta}}(\mathbf{x}) {E^L_{\bm{\theta}}(\mathbf{x})} \dr \mathbf{x},
}
where $H$ is the Hamiltonian, $\rho_{\bm{\theta}}(\mathbf{x}) = |\psi_{\bm{\theta}}(\mathbf{x})|^2$ is the probability density, and $E^L_{\bm{\theta}}(\mathbf{x}) = \frac{H 
\psi_{\bm{\theta}}(\mathbf{x})}{\psi_{\bm{\theta}}(\mathbf{x})}$ is called the \textit{local energy}. 
We have assumed that $\psi_{\bm{\theta}}(\mathbf{x})$ is square-normalized, which is always satisfied with the Waveflow ansatz. 
The final form is interpreted as the expectation value of $E^L_{\bm{\theta}}(\mathbf{x})$ under the distribution $\rho_{\bm{\theta}}(\mathbf{x})$, and can thus be approximated by Monte Carlo integration. 
In the VQMC approach, sampling is performed using Markov chains, resulting in Markov chain Monte Carlo (MCMC).
However, MCMC can often lead to biased sampling, 
while using normalizing flows allows us to draw exact samples from the distribution.

Let $E_{\rm exact}$ be the exact ground-state energy. According to the variational principle, $E_{\rm exact}\leq E(\bm{\theta})$. 
Therefore, the global minimum of $E(\bm{\theta})$ is the best approximation of $E_{\rm exact}$ under the ansatz $\psi_{\bm{\theta}}(\mathbf{x})$. 
We define the ground-state energy corresponding to $\psi_{\bm{\theta}}(\mathbf{x})$ as 
\eqsplit{\label{eq:vqmctarget}
E_{0} = \min_{\bm{\theta}} E(\bm{\theta}). 
}

The above minimization requires the evaluation of the gradient 
\eqsplit{
\frac{\partial E(\bm{\theta})}{\partial \theta}  =
\E_{\rho_{\bm{\theta}}(\mathbf{x}) }\left[ \frac{\partial \log(\rho_{\bm{\theta}}(\mathbf{x}))}{\partial \theta}  E^L_{\bm{\theta}}(\mathbf{x})\ + \frac{\partial E^L_{\bm{\theta}}(\mathbf{x})}{\partial \theta}  \right],
}
which can also be evaluated with sampling.

\begin{figure}[t!]
\centering
      \centering
      \includegraphics[width=0.9\linewidth]{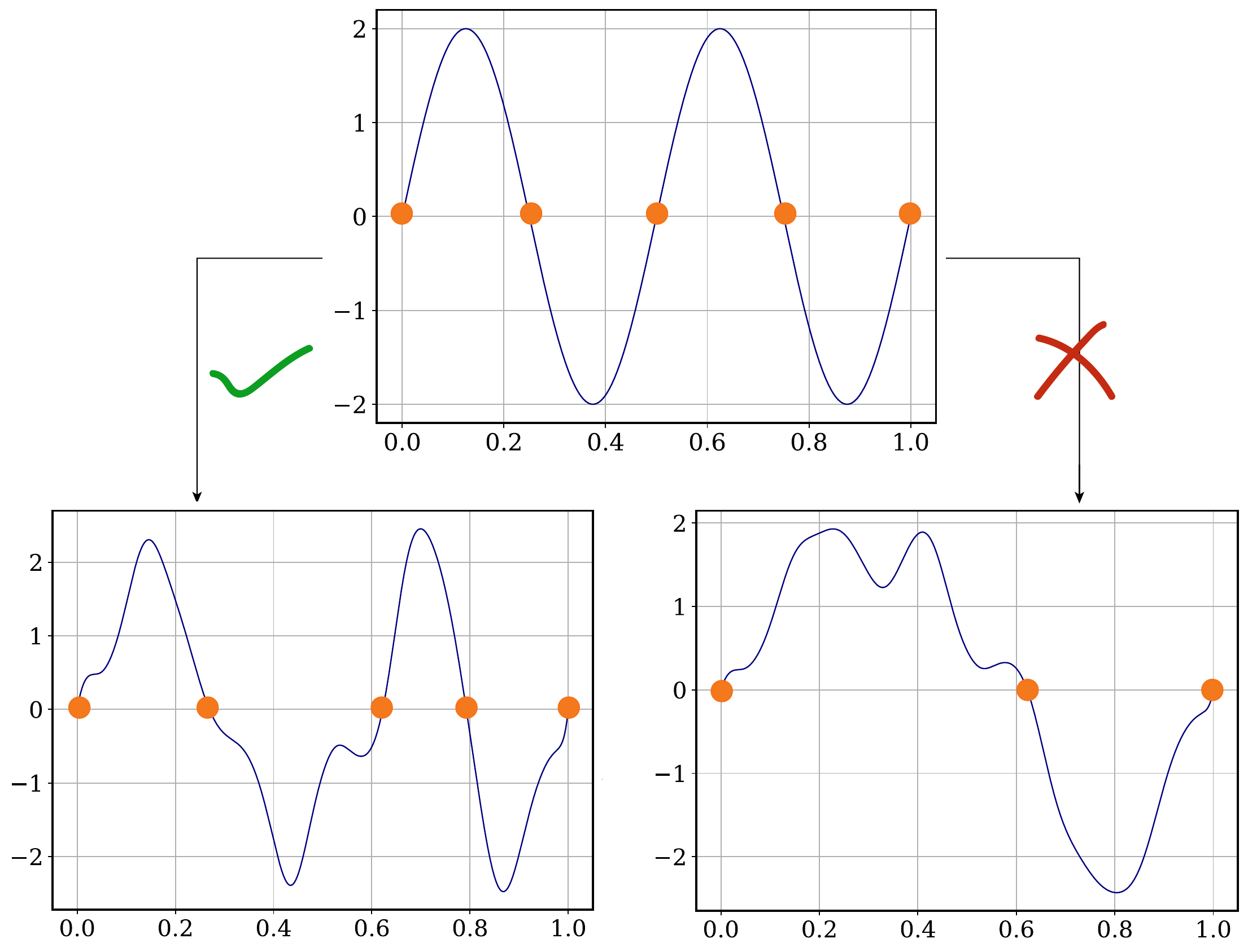}
      \caption{An example illustrating the node number mismatch between prior and target distributions.}
      \label{fig:node_number_mismatch}
\end{figure}

\subsection{Normalizing flows\label{sec:method_normalizing_flows}}
Normalizing flows transform a prior distribution $p_\mathbf{z}(\mathbf{z}):  \R^n \rightarrow \R$ into a new distribution $p_{\mathbf{x}}(\mathbf{x}):  \R^n \rightarrow \R$ via a learned continuous coordinate transform. 
The prior distribution $p_\mathbf{z}(\mathbf{z})$ is usually chosen to be simple, such as a Gaussian or uniform distribution. 
The diffeomorphic coordinate transformation $g: \R^n \rightarrow \R^n $  is defined as $\mathbf{x} = g(\mathbf{z})$. Therefore, $p_{\mathbf{x}}(\mathbf{x})$ can be derived by
\begin{align}\label{eq:normalizingflow}
    p_{\mathbf{x}}(\mathbf{x}) = p_{\mathbf{z}}(g^{-1}(\mathbf{x})) \cdot |\text{det}{ \, \jacobian{g^{-1}(\mathbf{x})}}|,
\end{align}
where $g^{-1}$ is the inverse of $g$, often referred to as the \textit{flow}, and $\jacobian{g^{-1}(\mathbf{x})}$ is the Jacobian matrix of $g^{-1}(\mathbf{x})$.

It is important to choose a proper form of the inverse coordinate transformation $g^{-1}$ such that the evaluation of the Jacobian is tractable.
In this work, we adopt autoregressive flows\cite{dinh2015nice, kingma2016improved}, 
where Eq.~\eqref{eq:normalizingflow} takes the form
\eqsplit{\label{eq:autoregressiveNormalizingFlow}
 p_{\mathbf{x}}(\mathbf{x}) &= \prod_{i=0}^{n-1} p_{z}(g^{-1}(x_i|\bm{\theta}_{\text{B},i})) \cdot \left\vert\partial_{x_i}  g^{-1}(x_i|\bm{\theta}_{\text{B},i})\right\vert, 
}
where $\bm{\theta}_{\text{B},i} = \bm{\theta}_{\text{B}}(\mathbf{x}_{0:i-1})$ are parameters of the bijection, and $i = 0, \ldots, n-1$ denotes each dimension of $\mathbf{x}$. 
$\bm{\theta}_{\text{B}}(\mathbf{x}_{0:i-1})$ indicates that the parameters for $z_i$ only depend on $(x_0, \ldots, x_{i-1})$.
$\bm{\theta}_{\text{B}}$ can be predicted by a neural network, called the conditioner.
Here, we choose the masked autoencoder for distribution
estimation (MADE)~\cite{germain2015made, papamakarios2017masked} as the neural network architecture to predict $\bm{\theta}_{\text{B}}$. 
Each bijection is composed of $L$ layers: 
$g^{-1}(\cdot|\bm{\theta}_{\text{B},i}) = g^{-1}(\cdot|\bm{\theta}^L_{\text{B},i}) \circ \ldots  \circ g^{-1}(\cdot|\bm{\theta}^1_{\text{B},i})$, 
where 
$\bm{\theta}^l_{\text{B},i} = \bm{\theta}_{\text{B}}^l(\mathbf{x}_{0:i-1})$ are the parameters of the $l$th layer. 
We choose $\bm{\theta}^l_{\text{B}, 0}$ to be a constant vector. 
Since $x_i$ only depends on the dimensions $\mathbf{x}_{0:i-1}$, the Jacobian $\jacobian{g^{-1}(\mathbf{x})}$ is a lower triangular matrix, whose determinant is simply the product of the diagonal elements. In Appendix~\ref{sec:apdx_normalizing_flows_algorithm}, we provide the algorithms for evaluating and sampling with normalizing flows.

Equivalently, we can define a square-normalizing flow:
\eqsplit{\label{eq:square_norm_anf}
\psi_{\mathbf{x}}(\mathbf{x}) =&\ \psi_{\mathbf{z}}(g^{-1}(\mathbf{x})) \cdot \sqrt{|\text{det}{ \,\jacobian{g^{-1}(\mathbf{x})}}|}\\
    =&\prod_{i=0}^{n-1} \psi_{z}(g^{-1}(x_i|\bm{\theta}_{\text{B},i})) \cdot \sqrt{|\partial_{x_i} g^{-1}(x_i|\bm{\theta}_{\text{B},i})|},
}
where we used $\psi_{\mathbf{x}}(\mathbf{x})$ to denote a square-normalized distribution, i.e., $\int |\psi(\mathbf{x})|^2 \mathrm{d}\mathbf{x} = 1$. 
For the purpose of learning an electronic wavefunction, one should use the square-normalizing flow. 
\begin{figure}[t!]
\centering        \includegraphics[width=0.9\linewidth]{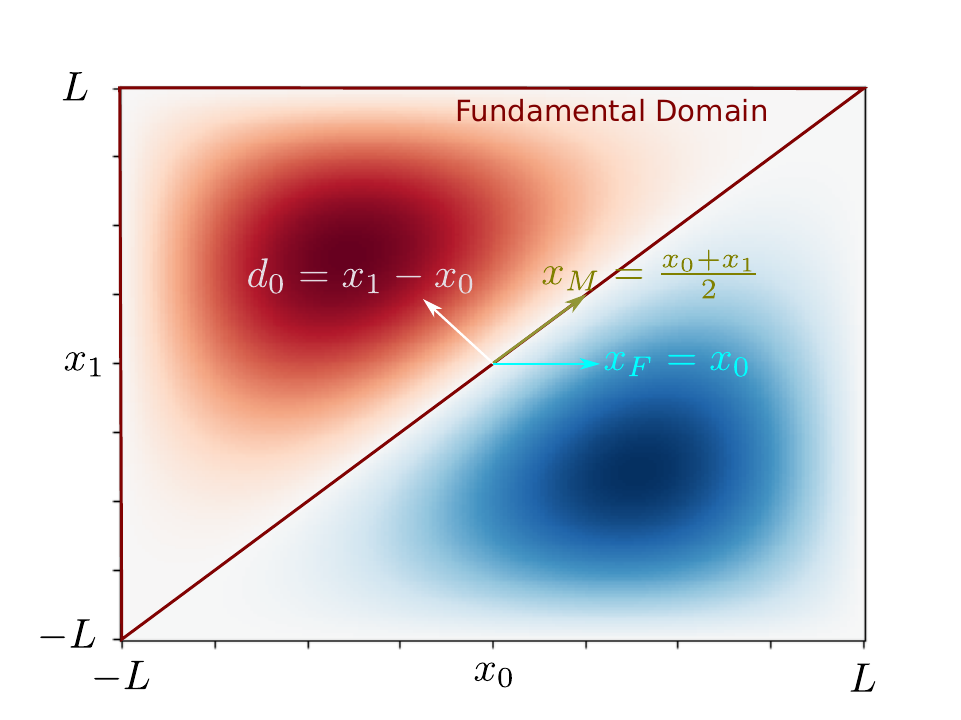}
    \caption{The wavefunction of two particles in a box, including the relative coordinate systems and the fundamental domain. The blue and red regions are equivalent, with either capable of serving as the fundamental domain.}
    \label{fig:two_particle_visualized}
\end{figure}

\begin{figure*}[t!]
    \centering
    \includegraphics[width=0.7\linewidth]{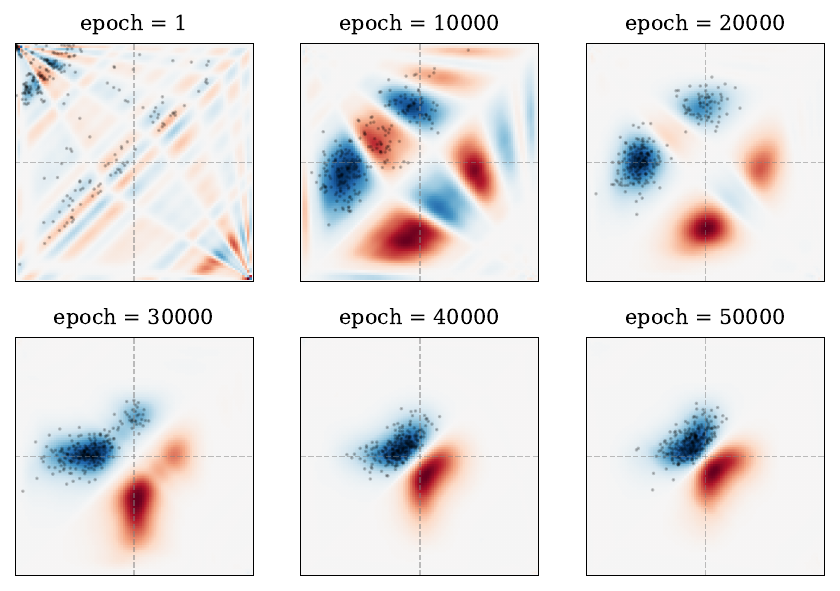}
\caption{Training progress of Waveflow for the ground state of a one-dimensional helium-like system. 
The blue region corresponds to the wavefunction with positive values and the red region with negative values. 
The $x$-axis and $y$-axis denote $x_0$ and $x_1$, respectively. 
The dots indicate samples drawn from the fundamental domain, chosen as the region where $d_1 = x_1 - x_0 \geq 0$.
}
\label{fig:training_progression}
\end{figure*}
\subsubsection{Bijections and priors\label{sec:spline_bijection_prior}}

The key components of normalizing flows are the priors and bijections, denoted as $p_\mathbf{z}(\mathbf{z})$ and $g^{-1}$ in Eq.~\eqref{eq:normalizingflow}. 
A good prior distribution should be simple and easy to sample from.
The bijection must be invertible and differentiable, with an easy-to-compute Jacobian determinant. 
Monotonic splines, such as the rational quadratic spline (RQS)~\cite{neural_splines}, provide a flexible yet controlled choice for bijections. 
However, RQS is only once differentiable, while higher-order differentiability is preferred for better trainability and stability. 
In particular, learning an electronic wavefunction requires the bijection to be at least twice differentiable because of the Laplacian operator in the Hamiltonian. 
Bijections with improved smoothness include mixture of logistic transformations~\cite{ho2019flow++}, deep (dense) sigmoidalflows~\cite{huang2018neural} and smooth bump functions~\cite{kohler2021smooth}.

In this work, we propose combining I-splines as the bijection and M/O-splines as the prior~\cite{ramsay1988monotone}. 
M-splines serve as the normalized prior, while O-splines are used for the square-normalized case.
The spline curves are constructed as linear combinations of polynomial basis functions. 
The weights of the basis functions serve as the parameters of the prior and bijection, denoted $\bm{\theta}_{\text{P}}$ and $\bm{\theta}_{\text{B}}$, respectively.
This setting allows precise control over the values and derivatives of the splines at given points, making it convenient to enforce boundary conditions. 
Moreover, using M/O-splines as the prior enhances flexibility, enabling the model to learn arbitrarily complex distributions and, therefore, overcome the topology mismatch problem discussed in the next subsection.

One drawback of the I-spline bijection is the lack of an analytical inverse. 
In this work, we use binary search to find its inverse up to a small numerical error. 
Since calling the I-spline function is computationally inexpensive 
compared to neural network calls, inverting the I-spline does not introduce a significant computational overhead.
However, the numerical inversion can become unstable when the curve is flat. 
To mitigate this, we regularize the minimal derivative by adding a small constant $\epsilon_\text{r}$ to all $\bm{\theta}_{\text{B}}$. 
The mathematical formulas of the splines mentioned above can be found in Appendix~\ref{sec:apdx_splines}.

\subsection{Topological mismatch\label{sec:topological_problem}}
The topological mismatch problem in normalizing flows arises when the target distribution has a different topological structure compared to the prior distribution. 
When this happens, it becomes impossible to perfectly match $p_{\mathbf{x},\text{learned}}(\mathbf{x})$ to $p_{\mathbf{x}, \text{target}}(\mathbf{x})$.  
For example, a Gaussian prior cannot be transformed into a distribution with two disconnected components.
Moreover, this can lead to numerical issues when inverting the bijection~\cite{cornish2020relaxing}. 
We illustrate the connectivity mismatch problem by attempting to reproduce the double-circle distribution with normalizing flows. 
In Fig.~\ref{fig:topological_two_circles_a}, we sampled $20,000$ points from the {scikit-learn}~\cite{scikit-learn} double-circle dataset with a Gaussian noise of width $0.05$. 
We used three affine coupling layers~\cite{dinh2015nice} to transform a two-dimensional Gaussian distribution into the target double-circle distribution. 
We then sampled $20,000$ points from the output of the $10,000$th epoch, as shown in Fig.~\ref{fig:topological_two_circles_b}.
Since the Gaussian distribution is connected everywhere, the affine normalizing flows failed to reproduce a disconnected distribution. 

The topological problem becomes more severe for the square-normalizing flows, where both positive and negative values are allowed. 
In addition to the connectivity mismatch, the number of nodes (points where the function changes sign) is also fixed during a continuous coordinate transformation. 
We illustrate the node number mismatch in Fig.~\ref{fig:node_number_mismatch}.
Starting from a one-dimensional square-normalized distribution between $[0, 1]$ with $5$ nodes shown in the upper panel, one can use square-normalizing flows to transform it into another distribution with $5$ nodes shown in the lower left panel. 
However, transforming this distribution to a distribution with a different number of nodes (e.g., $3$ nodes in the lower right panel) is impossible.
This node number mismatch makes learning an electronic wavefunction challenging because the number of nodes for the desired electronic wavefunction is usually unknown \textit{a priori}. 

The above topological problems can be tackled by introducing shape parameters $\bm{\theta}_{\text{P}}$ to the prior distribution. 
In autoregressive flows, these parameters can depend on the previous output, and the flows become
\eqsplit{\label{eq:anf_shape_params}
    p_{\mathbf{x}}(\mathbf{x}) = \prod_{i=0}^{n-1} p_{z}(g^{-1}(x_i|\bm{\theta}_{\text{B},i})|\bm{\theta}_{\text{P},i}) \cdot 
    \partial_{x_i} g^{-1}(x_i|\bm{\theta}_{\text{B},i}), 
}
where $\bm{\theta}_{\text{P},i} = \bm{\theta}_{\text{P}}(\mathbf{z}_{0:i-1})$ are the parameters for the prior. In this work, we choose the same MADE network as for the bijections to predict the shape parameters $\bm{\theta}_{\text{P}}$.
The same trick can be applied to the square-normalizing flow. 
Therefore, we choose a slightly more complicated while more flexible prior $p_\mathbf{z}(\mathbf{z}|\bm{\theta}_{\text{P}}(\mathbf{z}))$ and $\psi_\mathbf{z}(\mathbf{z}|\bm{\theta}_{\text{P}}(\mathbf{z}))$, while using the coordinate transformations $g^{-1}$ to perform the fine-tuning. 
As an example, in Fig.~\ref{fig:topological_two_circles_c}, we 
used Waveflow to reproduce the double-circle distribution. 
The Waveflow was constructed by a degree-$5$ M-spline curve with $23$ knots as the prior, and three layers of degree-$5$ I-spline curves with $23$ knots as the bijection.
We sampled $20,000$ points at the $10,000$th epoch. The regularization constant $\epsilon_\text{r}$ is set to be $0.02$.
The learned distribution successfully captured the target distribution's topology, which was not possible in the previous attempt shown in Fig.~\ref{fig:topological_two_circles_b}.
Normalizing flows algorithms with shape parameters are provided in 
Appendix~\ref{sec:apdx_normalizing_flows_algorithm}.

\subsection{Boundary conditions\label{sec:boundary_conditions}}

We consider the following $k$-th order boundary condition
\begin{align}
    \partial^k_{x_i} p_{\mathbf{x}}(\mathbf{x})|_{x_i=x^*} &= 0, \label{eq:bc3} 
\end{align}
where $k = 0, 1, \ldots$ and  $x^*$ is called the enforcement point.  In this work, we focus on the zeroth-order boundary condition. 

\begin{theorem}\label{thm:enforce_bc}
Let $g^{-1}_{l}(\mathbf{x})$ be a composition of autoregressive bijections with parameters $\bm{\theta}_{\mathrm{B}}^l$.
Assume that $g^{-1}_l(x_i = x^*|\bm{\theta}_{\mathrm{B},i}^l) = x^*$, for all $l$ and $\bm{\theta}_{\mathrm{B}}$, and that $p_\mathbf{z}(\mathbf{z})$ is an autoregressive prior distribution with $p_z(z_i=x^*|\bm{\theta}_{\mathrm{P},i}) = 0, \forall\bm{\theta}_{\mathrm{P}}$. Then any (square) normalizing flow defined in Eq.~\eqref{eq:anf_shape_params} fulfills $p_\mathbf{x}(\mathbf{x})|_{x_i=x^*} = 0$. 
\end{theorem}

\begin{proof}
Consider $\mathbf{x}$ with $x_i = x^*$, and let $g^{-1}$ be a bijection that maps $\mathbf{x}$ to a new point $\mathbf{x}'$ such that ${x}'_i = x^*$. The transformation of the $i$th dimension can be expressed as
\begin{equation*}
    g^{-1}(x_i=x^*|\bm{\theta}_{\text{B},i}) 
    = g^{-1}(x_i^L|\bm{\theta}_{\text{B},i}^L) \circ \ldots  \circ g^{-1}(x_i^1 = x^*|\bm{\theta}_{\text{B},i}^1)) = x^*,
\end{equation*}
where $x_i^{l}$ indicates the transformed $x_i$ after $l$ layers. Therefore,
\begin{align*}
     &p_\mathbf{x}(x_0,\ldots ,x_i=x^*,\ldots ,x_{n-1}) \\
     &=\ p_{z}(g^{-1}(x_0))\cdot \ldots \cdot p_{z}(g^{-1}(x_i = x^*|\bm{\theta}_{\text{B},i}))\cdot \ldots  \\   
     & \quad \cdot \partial_{x_0} g^{-1}(x_0)\cdot \ldots  \cdot 
     \partial_{x_i} g^{-1}(x_i=x^*|\bm{\theta}_{\text{B},i})\cdot \ldots \\
     &=\ p_{z}(g^{-1}(x_0))\cdot \ldots \cdot \underbrace{p_{z}(z_i = x^*)}_{=0}\cdot \ldots   \\ &\quad\cdot\partial_{x_0} g^{-1}(x_0)\cdot \ldots  \cdot  \partial_{x_i} g^{-1}(x_i=x^*|\bm{\theta}_{\text{B},i})\cdot \ldots  \\
     &=\ 0.
\end{align*}
\end{proof}

The preconditions of Theorem~\ref{thm:enforce_bc} can be met by ensuring the M/O/I-splines have the desired values at $x^*$. Similarly, the normalizing flows defined in Eq.~\eqref{eq:anf_shape_params} can also be adapted to enforce higher-order boundary conditions, which are not covered here for brevity.

\subsubsection{Antisymmetry\label{sec:antisymmetry}}
We begin by introducing two key concepts from group theory that are essential for enforcing antisymmetry to the Waveflow ansatz.
\begin{definition}[Orbit]
Let $G$ be a group that acts on a space $X$. The orbit of a point $x \in X$ is defined as
\eqsplit{
\mathrm{Orb}_G(x) := \{g \cdot x| g \in G \}.
}
\end{definition}
\begin{definition}[Fundamental domain]
Let $G$ be a group that acts on a space $X$, and let $S \subset X$ be a closed subset of $X$. Then we call $S$ the fundamental domain of $G$ if for all $x,y \in S$,
\begin{align}
    X = \cup_{x \in S} \mathrm{Orb}_G(x)  \ \mathrm{ and } \ \emptyset = \mathrm{Orb}_G(x) \cap \mathrm{Orb}_G(y).
\end{align}
\end{definition}
The fundamental domain $S$ serves as a minimal, non-redundant building block that covers the entire space without overlaps when repeated under the symmetry operators in $G$. 
Similarly, we can define the antisymmetric fundamental domain of a space under permutation operators. 
We only need to approximate the wavefunction within the fundamental domain and enforce necessary boundary conditions~\cite{klimyk2007antisymmetric}. In the following, we articulate this idea for the one-dimensional system.

We define a group $\mathcal{P}$ of permutation operators $P_\sigma$ acting on the spatial coordinates $({x}_0, {x}_1,\ldots ,{x}_{n-1})$ such that
\begin{align}
    P_\sigma \cdot ({x}_0, {x}_1,\ldots ,{x}_{n-1}) = (\sigma({x}_0), \sigma({x}_1),\ldots , \sigma({x}_{n-1})),
\end{align}
where $\sigma$ represents a specific permutation. 
The fundamental domain of $\mathcal{P}$ is then given by: 
\begin{theorem}
$S = \{{x}_0 \leq {x}_1 \leq \ldots  \leq {x}_{n-1} | {x}_0,\ldots ,{x}_{n-1} \in \R\}$ is a fundamental domain of $\mathcal{P}$ on $\R^n$.
\end{theorem}

We define the antisymmetric wavefunction within the above fundamental domain $S$. Moreover, we impose the following boundary condition 
\begin{align}\label{eq:fdboundarycondition}
    \psi({x}) = 0, \quad \forall {x} \in \partial S, 
\end{align}
where $\partial S$ denotes the boundary of $S$, corresponding to where at least two coordinates coincide. 
Eq.~\ref{eq:fdboundarycondition} corresponds to the Pauli exclusion principle, meaning that two identical electrons cannot be found at the same position.

To illustrate the definition of fundamental domains, we consider the ground state of two electrons in a box ranging from $-L$ to $L$, analytically evaluated as
\eqsplit{ 
    \psi(x_0, x_1) = \frac{1}{\sqrt{2}L} &\left[ \cos{\left(\frac{\pi x_0}{2L}\right)}\sin{\left(\frac{\pi x_1}{L}\right)}\right. \\
    &- 
  \left.\cos{\left(\frac{\pi x_1}{2L}\right)}\sin{\left(\frac{\pi x_0}{L}\right)}\right].
}
We plot the ground state wavefunction in Fig.~\ref{fig:two_particle_visualized} (b). The fundamental domain can be chosen as either the blue or red regions.

\subsection{Waveflow algorithm\label{sec:waveflow_alg}}
This section presents the Waveflow algorithm, focusing on one-dimensional many-electron systems confined within $[-L, L]$.
This confinement applies to most one-dimensional systems since we can make $L$ arbitrarily large. 
We transform the absolute coordinates $\{x_0,\ldots ,x_{n-1}\}$ to relative coordinates $\{d_i = x_{i+1} - x_{i}| i \in \{0,\ldots ,n-2\}\}$ plus an extra unconstrained coordinate $x_u$ describing the global position of the particles. 
$x_u$ can be chosen as the average of the absolute coordinates, denoted as $x_M = \frac{1}{n}\sum_i x_i$, or the absolute coordinate of the first particle, denoted as $x_F=x_0$. 
The fundamental domain is given by  $\{d_0,\ldots ,d_{n-2}, x_u| 0\leq d_i \leq 2L, \forall i; -L\leq x_u\leq L\}$. 

We use I-splines as the bijection $g^{-1}$.
Since I-splines map $[0,1]$ to $[0,1]$, we rescale the relative coordinates into the $[0,1]$ range. 
For square-normalizing flows, we take the coordinate transformation and rescaling as the parameter-free first layer.
Antisymmetry is fulfilled by enforcing $\psi(d_0,\ldots, d_{n-2}, x_u) = 0$ when $\exists d_i = 0$. Moreover, we require the wavefunction to be $0$ at the wall of the box, corresponding to $\exists d_i = 1$ or $x_u = 0$ or $1$. Therefore, the boundary conditions of the square-normalizing flows in the fundamental domain are 
\eqsplit{
    \psi(d_0,\ldots,d_i=0,\ldots, d_{n-2}, x_u) &= 0,\ \forall i,\\
    \psi(d_0,\ldots,d_i=1,\ldots, d_{n-2}, x_u) &= 0,\ \forall i, \\
    \psi(d_0,\ldots , d_{n-2}, x_u=0) &= 0, \\
    \psi(d_0,\ldots , d_{n-2}, x_u=1) &= 0.
}
The above boundary conditions are enforced using Theorem~\ref{thm:enforce_bc}. Subsequently, the Waveflow ansatz is trained by VQMC described in Section~\ref{sec:vqmc}.
The overall complexity of the model scales as $\mathcal{O}\left(TN_{\text{s}}n^3\right)$, where $T$ is the number of gradient decent steps, $N_{\text{s}}$ is the number of samples, and $n$ is the number of electrons. A detailed complexity analysis is provided in Appendix~\ref{sec:apdx_complexity}.

Once the wavefunction is evaluated within the fundamental domain $S$, denoted as $\psi_S(d_0,\ldots,d_{n-2}, x_u)$, we can expand it to the whole space. 
We identify the permutation between points outside and inside $S$, and use $\eta(x_0, \ldots ,x_{n-1})$ to represent the number of pair-wise exchanges, then
\eqsplit{\label{eq:evaluate_waveflow_everywhere}
    \psi(x_0,\ldots ,x_{n-1}) = (-1)^{\eta(x_0,\ldots ,x_{n-1})}\psi_S(d_0,\ldots ,d_{n-2}, x_u).
}
Note that for the purpose of evaluating a physical observable, such as the energy, the wavefunction in the fundamental domain is sufficient. 
We expand the wavefunction to the whole space using Eq.~\eqref{eq:evaluate_waveflow_everywhere} only for visualization purposes.

\begin{figure*}
\centering
\begin{subfigure}[t]{0.3\textwidth}
  \includegraphics[width=1\linewidth]{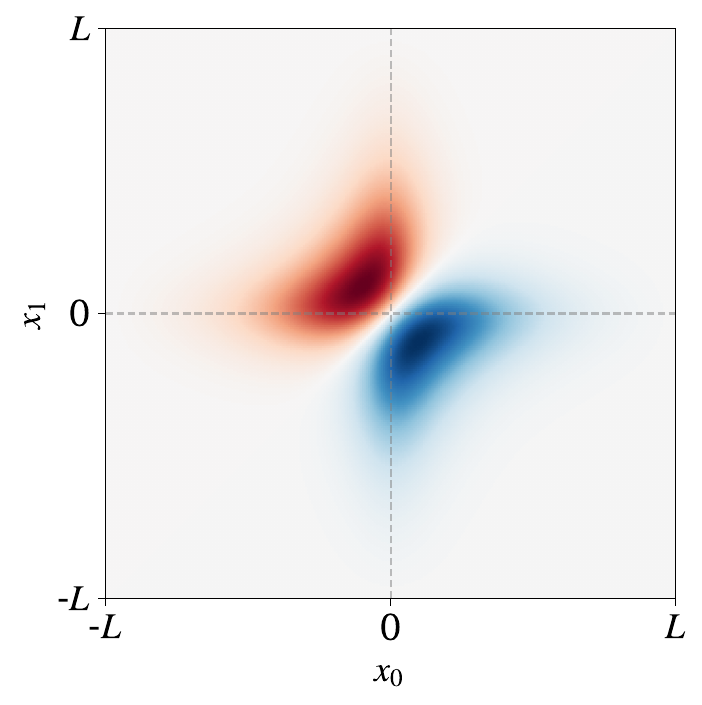}
  \caption{QMSolve}
  \label{fig:helium_ground_state_a}
\end{subfigure}
\begin{subfigure}[t]{0.3\textwidth}
  \includegraphics[width=1\linewidth]{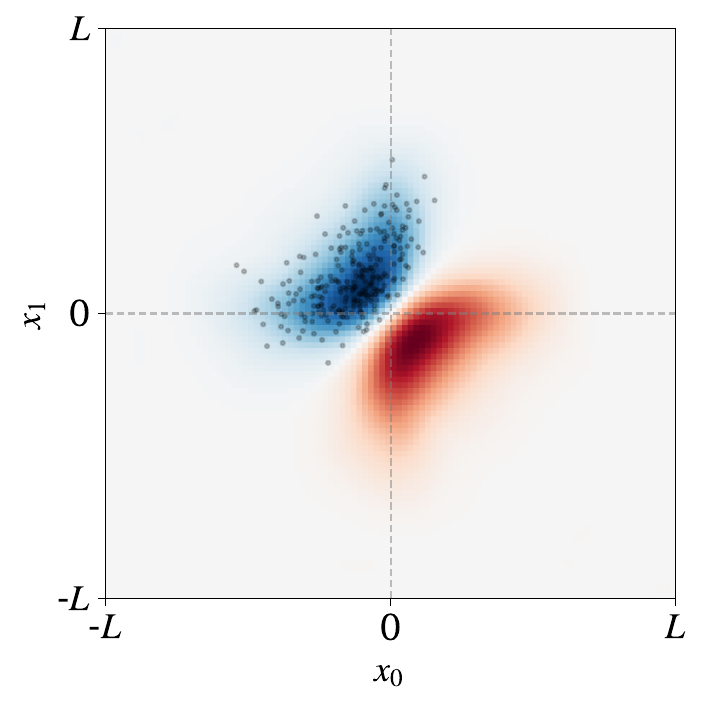}
  \caption{Waveflow}
  \label{fig:helium_ground_state_b}
\end{subfigure}
\begin{subfigure}[t]{0.35\textwidth}
  \includegraphics[width=1\linewidth]{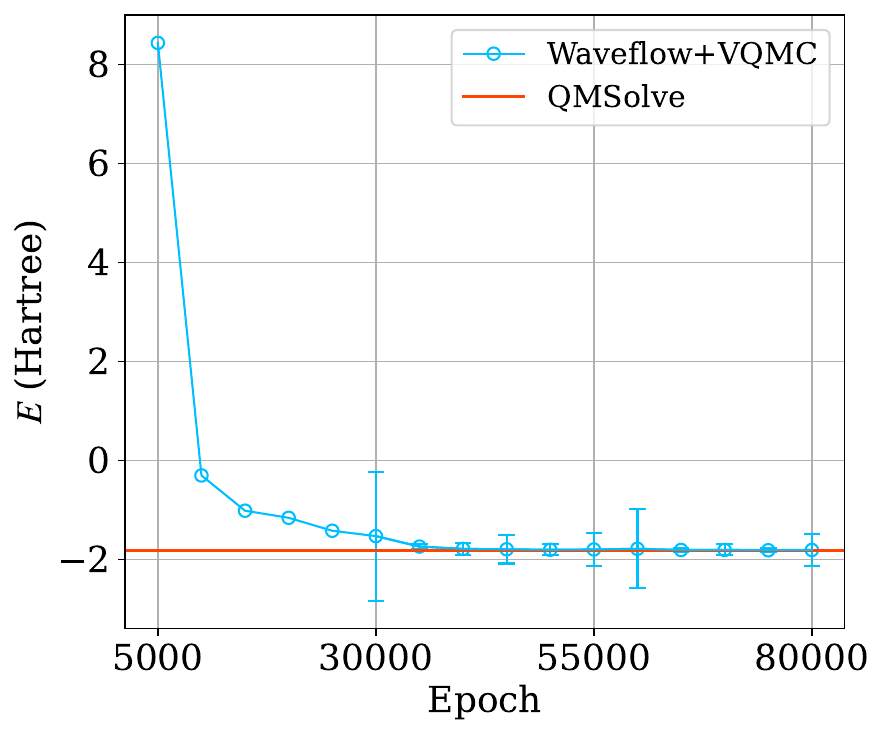}
  \caption{Energy convergence}
  \label{fig:helium_ground_state_c}
\end{subfigure}
\caption{
Evaluations of the ground-state wavefunction and energy for the one-dimensional helium-like system.
(a) Ground state wavefunction evaluated by QMSolve.
(b) Ground state wavefunction evaluated by Waveflow.
(c) Convergence of the Waveflow energy with respect to the number of epochs.
}
\end{figure*}

\section{Experiments}\label{sec:experiments}
We study a one-dimensional spinless electronic system with a central nuclear-like potential within $[-L, L]$ and an infinite potential for regions outside this interval.
The Hamiltonian involves two electrons, making the ground state straightforward to visualize.  
We set the "nuclear" charge to be $2$ and refer to this toy model as a one-dimensional helium-like system. 
The Hamiltonian is defined as
\eqsplit{\label{eq:helium_hamiltonian}
    H = - \frac{1}{2} \sum_{i=0,1} \nabla_i^2 - \sum_{i=0,1}\frac{2}{\sqrt{1 + x_i^2}} + \frac{1}{\sqrt{1 + (x_0-x_1)^2}},
}
where we used the soft Coulomb potential for both electron-nuclear and electron-electron interactions, with the softening parameter equal to $1$. 
The soft Coulomb potential avoids dealing with cusp conditions, which is beyond the scope of this work. 
All units are in the atomic units (a.u.), with the box length $L = 10$ Bohr. 

The Waveflow ansatz was constructed with a degree-$5$ O-spline curve prior with $23$ knots, along with a bijection comprising three layers of degree-5 I-spline curves, each with 23 knots. The degree of the splines determines their differentiability, while the number of knots controls their flexibility.
For calculating the Laplacian, the curves must be at least twice differentiable, which requires I-spline curves of at least degree 2 and O-spline curves of at least degree 3. However, to represent the ground state for a Hamiltonian with a smooth potential, higher differentiability is desired.
On the other hand, increasing the degree of the splines can reduce their flexibility.
Therefore, we need a larger number of knots to maintain flexibility, which, unfortunately, increases computational cost. 
We found that our chosen configuration strikes a good balance between smoothness, flexibility, and computational efficiency.
The regularization constant $\epsilon_\text{r}$ for the bijection is set to be $0.05$. 
The unconstrained coordinate $x_u$ is defined as $x_u = x_M = (x_0 + x_1)/2$.
We optimize the network using the {Adam} algorithm~\cite{kingma2014adam} with a learning rate $\eta = 10^{-4}$, and exponential decay rates for the first and second moment estimates set to $\beta_1 = 0.9$ and $\beta_2=0.999$. 
The MADE network has one hidden layer of $64$ neurons.

Fig.~\ref{fig:training_progression} shows the training progress of the Waveflow. 
The $x$-axis and $y$-axis represent the coordinates of each electron. 
Starting from a prior distribution with an arbitrary number of nodes at epoch $1$, the wavefunction was transformed to minimize the ground-state energy estimated by VQMC, with 
$256$ points sampled from the fundamental domain.
The sampled points are shown as dark dots in the plot.
As the training progressed, the number of nodes decreased and eventually converged to the correct number of the ground state.
This result demonstrates that Waveflow can overcome the topological mismatch between the prior and the target distributions. 
We chose the solution from the QMSolve package~\cite{qmsolve} as the reference to benchmark our results.
We compared the wavefunction derived from Waveflow in Fig.~\ref{fig:helium_ground_state_b} to that from QMSolve in Fig.~\ref{fig:helium_ground_state_a}. 
The comparison shows excellent agreement between the two solutions, with a global phase difference $-1$, which does not affect the evaluation of physical observables.
Fig.~\ref{fig:helium_ground_state_c} shows the convergence of the ground-state energy with respect to the training epochs, which averaged the energy value every $5,000$ epochs.
We plotted the error bars after the $30,000$th epoch due to the large error bars in the earlier steps.
The best ground-state energy estimated by Waveflow was $-1.8123 \pm 0.038$ Hartree, compared to the $-1.8125$ Hartree from QMSolve.
The results indicate that our approach is able to recover the correct nodal structure of the ground-state wavefunction, while converging to the accurate ground-state energy.
The converged Waveflow model represents a combination of prior and flow that leads to the correct target distribution, where the prior focuses on approximating the nodal structure and the flow fine-tunes the prior into the correct ground-state wavefunction.

\section{Conclusions\label{sec:conclusions}}
In this work, we introduced Waveflow, an NNQS framework that directly learns an electronic wavefunction with boundary-conditioned normalizing flows. 
By defining the fundamental domain of the wavefunction and applying necessary boundary conditions, Waveflow imposes antisymmetry without relying on Slater determinants, offering a potentially more expressive ansatz for complex many-electron systems. 
Moreover, our approach addressed the topological mismatch between prior and target distributions by using O-spline priors and I-spline bijections, which enabled flexibility in the node number of the distribution while maintaining its square-normalization property.
Our framework is readily applicable to other distribution learning tasks that require specific boundary conditions or suffer from topological mismatches, such as connectivity mismatch for normalizing flows and node number mismatch for square-normalizing flows. 
Furthermore, this ansatz can be extended beyond electronic structures. For instance, its automatic square normalization feature positions it as a promising candidate for applications in quantum dynamics simulations.
Nevertheless, the current stage of Waveflow is still proof-of-concept, and a more efficient implementation is desired to handle large-scale experiments.
In summary, Waveflow represents a significant step forward in the application of normalizing flows to quantum chemistry. While there are challenges to overcome, the potential benefits of this approach warrant further exploration and development.

\section*{Data availability statement}
The Waveflow code and experiments can be found at \href{https://github.com/aspuru-guzik-group/waveflow}{https://github.com/aspuru-guzik-group/waveflow}.

\section*{Acknowledgement}
This work was supported in part by the Acceleration
Consortium and the Vector Institute.
A.A.-G. also acknowledges the generous support from Dr. Anders G. Fr{\o}seth, Natural Resources Canada, and the Canada 150 Research Chairs
Program.
\appendix
\setcounter{algocf}{0}
\renewcommand{\thealgocf}{A\arabic{algocf}}

\section{Normalizing flow algorithmss\label{sec:apdx_normalizing_flows_algorithm}}
We provide the pseudocodes for evaluating and sampling from an autoregressive normalizing flow. 
These algorithms can be adapted for square-normalizing flows by replacing the normalized prior distribution $p_\mathbf{z}(\mathbf{z})$  with square normalized $\psi_\mathbf{z}(\mathbf{z})$, and substituting the determinant of the Jacobian with the square root of the determinant, as summarized in Eq.~\eqref{eq:square_norm_anf}.
Additionally, for square-normalizing flows, sampling is based on $|\psi_\mathbf{z}(\mathbf{z})|^2$ instead of $p_\mathbf{z}(\mathbf{z})$. 

The algorithms for standard normalizing flows are presented in Algorithms~\ref{alg:normalizingfloweval} and \ref{alg:normalizingflowsample}. 
The algorithms with adaptive prior described in this work are summarized in Algrithms~\ref{alg:normalizingflowevalAdaptivePrior} and \ref{alg:normalizingflowsampleAdaptivePrior}.
We use $\bm{\theta}_{\text{NN}}$ for the parameters of the neural networks, $\bm{\theta}_{\text{B}}^l$ for the parameters of the $l$th layer of the bijection, and $\bm{\theta}_{\text{P}}$ for the parameters of the prior.

\begin{algorithm}[h!]
\caption{Evaluating an autoregressive normalizing flow}
\label{alg:normalizingfloweval}
\SetAlgoLined
\KwIn{samples $\mathbf{x}$, prior  $p_{\mathbf{z}}(\mathbf{z})$, and NN parameters $\bm{\theta}_{\text{NN}}$}
\KwOut{$p_{\mathbf{x}}(\mathbf{x}) = p_{\mathbf{z}}(g^{-1}(\mathbf{x})) \cdot |\text{det}{ \, \jacobian{g^{-1}(\mathbf{x})}}|$}
Initialize $\text{det} \gets 1$ \\
\For{$l \gets 0$ \KwTo $L-1$}{
    $\bm{\theta}_{\text{B}}^l = \bm{\theta}_{\text{B}}^l(\mathbf{x})$
    \\
    \For{$i \gets 0$ \KwTo $n-1$}{
        $x^{l+1}_i = g^{-1}(x_i^l|\mathbf{\theta}_{\text{B},i}^l)$  \\
        $\text{det} = \text{det} \cdot \partial_x g^{-1}(x_i^l|\mathbf{\theta}_{\text{B},i}^l)$ \\
    }
}
\Return $p_{\mathbf{x}}(\mathbf{x}) = \prod_{i=0}^{n-1} p_z(x_i^L) \cdot \text{det}$
\end{algorithm}

\begin{algorithm}[h!]
\caption{Sampling an autoregressive normalizing flow}
\label{alg:normalizingflowsample}
\SetAlgoLined
\KwIn{NN parameters $\bm{\theta}_{\text{NN}}$}
\KwOut{$\mathbf{x}$ sampled from $p_\mathbf{x}(\mathbf{x})$}
Draw a sample from the prior $\mathbf{x}^L \sim p_\mathbf{z}(\mathbf{z})$ \\
\For{$l \gets L-1$ \KwTo $0$}{
    \For{$i \gets 0$ \KwTo $n-1$}{
        $\mathbf{\theta}^l_{\text{B},i} = \bm{\theta}^l_{\text{B}}(x^l_{0}, ..., x^l_{i-1})$ \\
        $x^l_i = g(x_i^{l-1}|\mathbf{\theta}^l_{\text{B},i})$ \\
    }
}
\Return $\mathbf{x}^0$
\end{algorithm}

\begin{algorithm}[h!]
\caption{Evaluating an autoregressive normalizing flow with adaptive prior}
\label{alg:normalizingflowevalAdaptivePrior}
\SetAlgoLined
\KwIn{samples $\mathbf{x}$, prior  $p_{\mathbf{z}}(\mathbf{z})$, and NN parameters $\bm{\theta}_{\text{NN}}$}
\KwOut{$p_{\mathbf{x}}(\mathbf{x}) = p_{\mathbf{z}}(g^{-1}(\mathbf{x})) \cdot |\text{det}{ \, \jacobian{g^{-1}(\mathbf{x})}}|$}
Initialize $\text{det} \gets 1$ \\
\For{$l \gets 0$ \KwTo $L-1$}{
   $\bm{\theta}_{\text{B}}^l = \bm{\theta}_{\text{B}}^l(\mathbf{x})$ \\
    \For{$i \gets 0$ \KwTo $n-1$}{
    $x^{l+1}_i = g^{-1}(x_i^l|\mathbf{\theta}_{\text{B},i}^l)$  \\
        $\text{det} = \text{det} \cdot \partial_x g^{-1}(x_i^l|\mathbf{\theta}_{\text{B},i}^l)$ \\
    }
}
$\bm{\theta}_{\text{P}} = \bm{\theta}_{\text{P}} (\mathbf{x})$ \\
\Return $p_{\mathbf{x}}(\mathbf{x}) = \prod_{i=0}^{n-1} p_z(x_i^L| \bm{\theta}_{\text{P}, i}) \cdot \text{det}$ 

\end{algorithm}

\begin{algorithm}[h!]
\caption{Sampling an autoregressive normalizing flow with adaptive prior}
\label{alg:normalizingflowsampleAdaptivePrior}
\SetAlgoLined
\KwIn{NN parameters $\bm{\theta}_{\text{NN}}$}
\KwOut{$\mathbf{x}$ sampled from $p_\mathbf{x}(\mathbf{x})$}
Draw a sample from the adaptive prior: 
$x_0^L \sim p_z(z|\bm{\theta}_{\text{P}, 0})$ \\
\For{$i \gets 0$ \KwTo $n-1$}{ 
    $\bm{\theta}_{\text{P}, i} =\bm{\theta}_{\text{P}}(x^L_{0}, ..., x^L_{i-1})$ \\
    $x^L_i \sim p_z(z| \bm{\theta}_{\text{P}, i})$ 
}
Transform the sample: \\
\For{$l \gets L-1$ \KwTo $0$}{
    \For{$i \gets 0$ \KwTo $n-1$}{
        $\mathbf{\theta}^l_{\text{B},i} = \bm{\theta}^l_{\text{B}}(x^l_{0}, ..., x^l_{i-1})$ \\
        $x^l_i = g(x_i^{l-1}|\mathbf{\theta}^l_{\text{B},i})$ \\
    }
}
\Return $\mathbf{x}^0$
\end{algorithm}

\section{Complexity Analysis\label{sec:apdx_complexity}}

In this section, we provide the complexity analysis of the Waveflow model. 
We first evaluate the cost of one model evaluation, $C_{\text{model}}$. 
Each model evaluation involves running the neural networks that encode electron positions, which scales as $\mathcal{O}(nM_h)$, where $n$ is the number of electrons and $M_h$ is the hidden dimension. 
Although $M_h$ is a hyperparameter, it is practical to choose $M_h$ to be proportional to the number of the electrons, $M_h \propto n$. Therefore, $C_{\text{model}}$ scales as $\mathcal{O}(n^2)$.

The training cost per sample per step can be divided into three components: sampling ($C_{\text{samp}}$), energy evaluation ($C_{\text{energy}}$), and gradient evaluation ($C_{\text{grad}}$).
For each sample, the number of model evaluations scales linearly with the number of electrons $n$, as we autoregressively sample each degree of freedom. 
Therefore, $C_{\text{samp}}$ scales as  $\mathcal{O}(n^3)$.
The dominant cost in energy evaluation comes from computing the Laplacian. 
For each sample, we calculate the second derivative for each electron. Using automatic differentiation, the cost of each second derivative is a constant multiple (approximately 5 times) of the cost of model evaluation. 
Therefore, $C_{\text{energy}}$ also scales as $\mathcal{O}(n^3)$, but with a larger computational overhead.
The cost of gradient evaluation, $C_{\text{grad}}$, remains constant, as automatic differentiation is used to compute the gradients of the neural network parameters.

Summing these components, the total cost of the Waveflow model scales as 
$\mathcal{O}\left(TN_{\text{s}}n^3\right)$, where $T$ is the number of gradient descent steps, and $N_\text{s}$ is the number of samples. 


\section{Splines\label{sec:apdx_splines}}
 In this section, we briefly review the different types of splines used in this work.

\subsection{B-splines}
\label{sec:B-splines}
\begin{figure}[t!]
\centering
\begin{subfigure}[t]{0.49\linewidth}
  \includegraphics[width=1\linewidth]{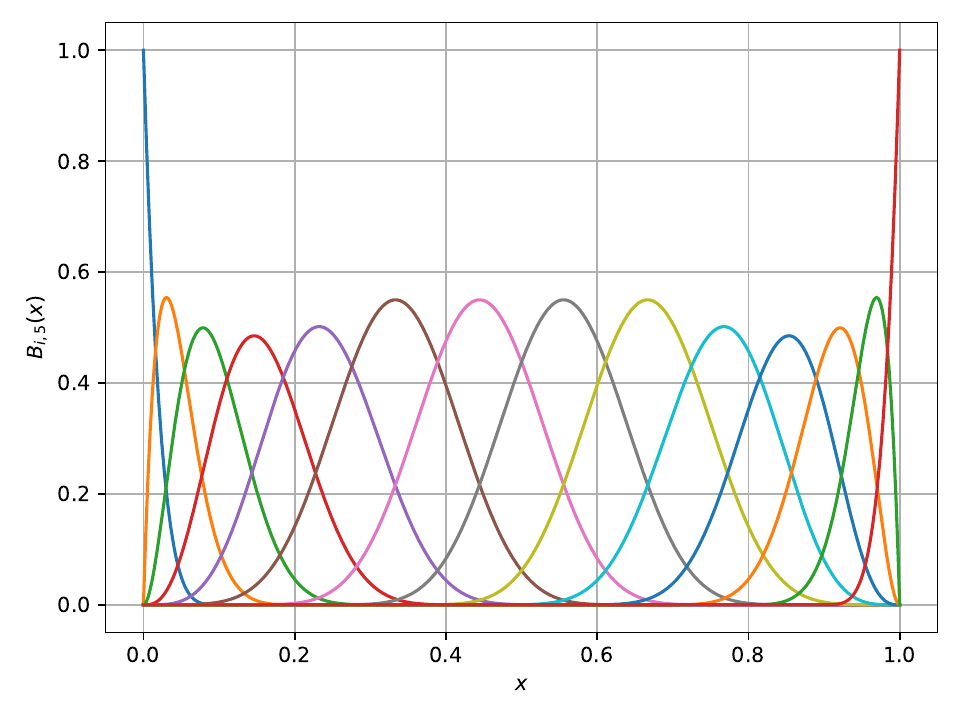}
  \caption{}
  \label{fig:bsplines}
\end{subfigure}
\begin{subfigure}[t]{0.49\linewidth}
    \includegraphics[width=1\columnwidth,]{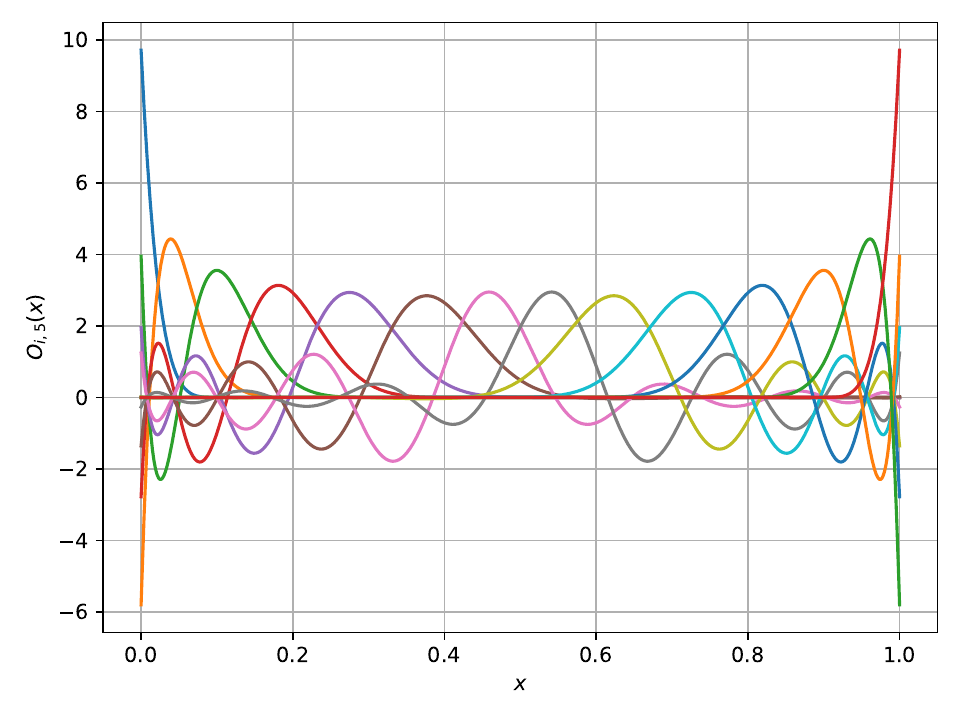}\hspace*{\fill}
    \caption{}
    \label{fig:obsplines}
 \end{subfigure}   \\
\begin{subfigure}[t]{0.49\linewidth}
  \includegraphics[width=1\linewidth]{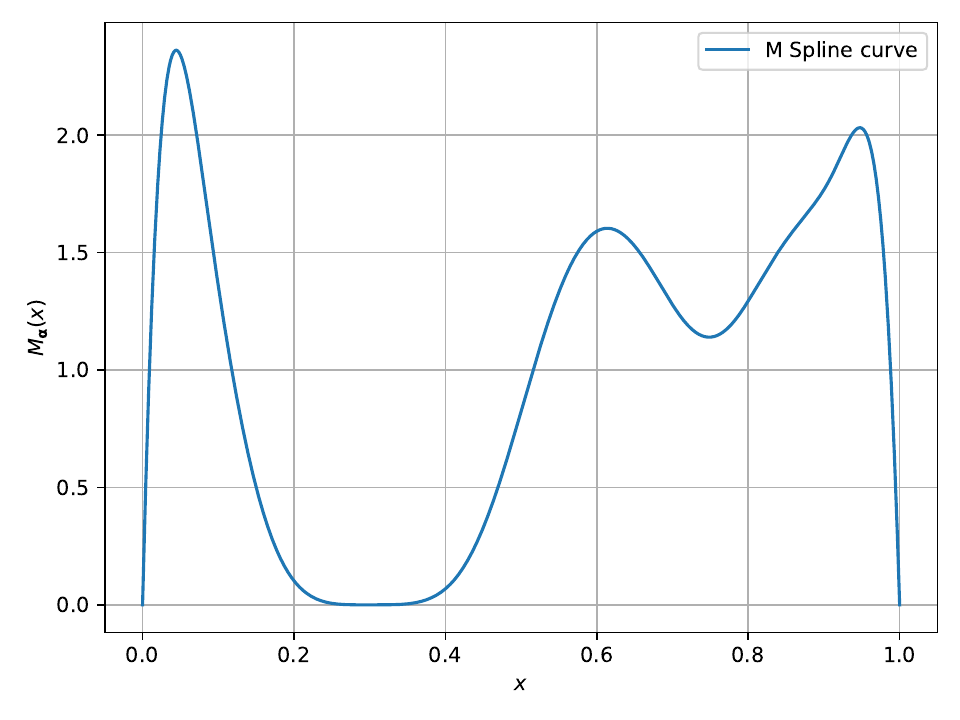}
  \caption{}
\label{fig:msplinecurve}
\end{subfigure}
\hfill
\begin{subfigure}[t]{0.49\linewidth}
\includegraphics[width=1\linewidth]{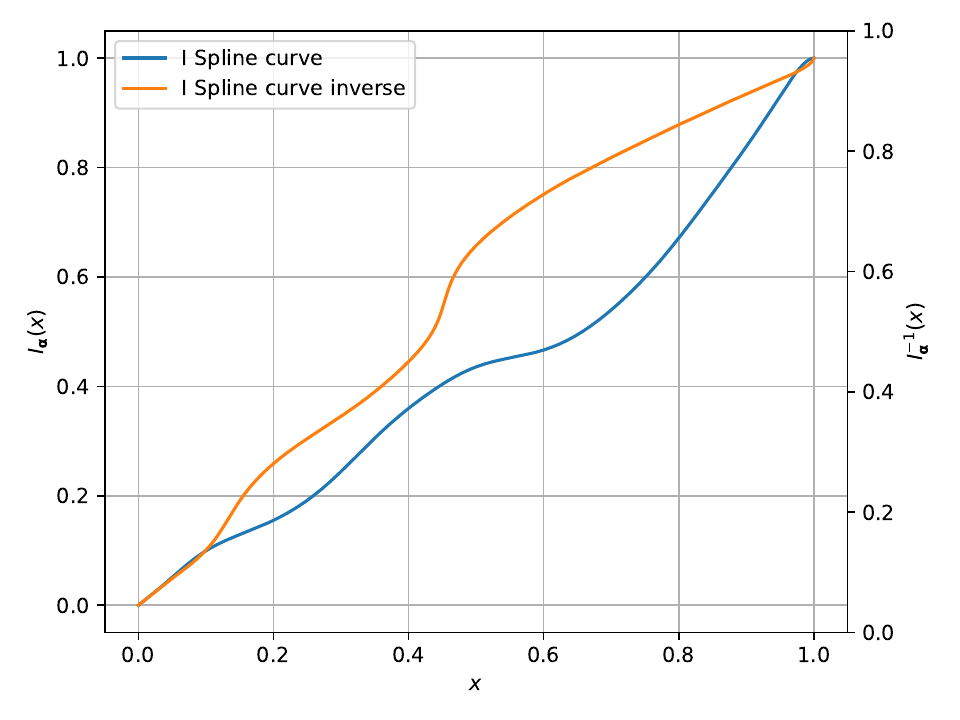}
  \caption{}
  \label{fig:isplinecurve}
\end{subfigure}
\caption{
Examples of splines and spline curves.
(a) 14 B-splines of degree $k=5$.
(b) O-splines from orthogonalization of (a).
(c) An M-spline curve.
(d) An I-spline curve and its inverse.
}
\label{fig:apdx_splines}
\end{figure}

The B-spline basis functions are defined recursively. Given a set of non-increasing knots $\{t_i\}$, the B-splines are defined as
\eqsplit{
B_{i, 1}(x|\mathbf{t}) =& \begin{cases}
    1, & \text{if } t_i < x < t_{i+1}, \\ 
    0, & \text{otherwise};
\end{cases}\\
B_{i,k}(x|\mathbf{t}) =& \frac{x - t_i}{t_{i+k} - t_i} B_{i, k-1}(x|\mathbf{t}) \\
&+ \frac{t_{i+k+1} - x}{t_{i+k+1} - t_{i+1}}B_{i+1, k-1}(x|\mathbf{t}), k > 1.
}
The B-splines satisfy that $B_{i,k}(x|\mathbf{t}) \geq 0$ and $\sum_i B_{i,k}(x|\mathbf{t}) = 1$. We call $k$ the degree of the splines. B-splines of degree-$5$ is shown in Fig.~\ref{fig:bsplines}.

\subsection{M-splines}
M-splines\cite{curry1988polya} are normalized B-splines satisfying $\int_{t_i}^{t_{i+1}} M_{i,k}(x|\mathbf{t}) \dx=1$ for any $i$ and $k$ values.  Thus has the form
\eqsplit{\label{eq:apdx_msplines}
M_{i,1}(x |\mathbf{t}) &= \begin{cases}(t_{i+1} - t_i)^{-1}, & t_i < x < t_{i+1}; \\
0, & \text{otherwise}.
\end{cases}\\
M_{i,k}(x |\mathbf{t}) &= \frac{k}{k-1}\left(\frac{x-t_i}{t_{i+k}-t_i}M_{i,k-1}(x |\mathbf{t}) \right. \\
&+ \left.\frac{t_{i+k}-x}{t_{i+k}-t_i} M_{i+1, k-1}(x |\mathbf{t})\right), k>1.
}
Next we define an M-spline curve $M_{k}(x |\mathbf{t}, \bm{\alpha})$ by summing up all $M_{i,k}(x |\mathbf{t})$ splines with weights $\{\alpha_i\}$,
\eqsplit{
M_k(x|\mathbf{t}, \bm{\alpha}) = \sum_{i=1}^N \alpha_i M_{i,k}(x|\mathbf{t}),
}
where  $\alpha_i>0$ and $\sum_{i}\alpha_i = 1$. Therefore $M_k(x|\mathbf{t}, \bm{\alpha})$ is normalized. We sample $M_k(x|\mathbf{t}, \bm{\alpha})$ with rejection sampling, which requires an upper bound for $M_k(x|\mathbf{t}, \bm{\alpha})$.

\begin{theorem}\label{theorem:maxM}
$M_k(x|\mathbf{t}, \bm{\alpha})$ is bounded by
\begin{align}\label{eq:apdx_msplines_max_bond}
    {M_k(x|\mathbf{t}, \bm{\alpha})} \leq k \max_{i, x}{M_{i,k}(x |\mathbf{t})}.
\end{align}
\end{theorem}

\begin{proof}
According to Eq.~\eqref{eq:apdx_msplines}, for any $x$ value, there are only $k$ $M_{i,k}(x)$ that are non-zero. Therefore, at $x=x_0$, we can find a number $j_0$ such that 
\eqsplit{
M_k(x_0|\mathbf{t}, \bm{\alpha}) &= \sum_{i=1}^N \alpha_i M_{i,k}(x_0|\mathbf{t}) = \sum_{i=j_0}^{j_0+k} \alpha_i M_{i,k}(x_0|\mathbf{t}) \\
}
In addition, since $0\leq \alpha_i \leq 1$, 
\eqsplit{
M_k(x_0|\mathbf{t}, \bm{\alpha}) &\leq \sum_{i=j_0}^{j_0+k} M_{i,k}(x_0|\mathbf{t})\\& \leq \sum_{i=j_0}^{j_0+k} \max_x{M_{i,k}(x|\mathbf{t})}
\leq k \max_{i,x}  M_{i,k}(x|\mathbf{t}). 
}
The above derivation applies to any $x_0$ values. Therefore, Eq.~\eqref{eq:apdx_msplines_max_bond} holds.
\end{proof}

We only sample one dimension at a time, making the rejection sampling efficient. Once the partition $\mathbf{t}$ is chosen, the double max in Eq.~\eqref{eq:apdx_msplines_max_bond} can be evaluated numerically by precomputing the basis once before training. An example of M-spline curves is shown in Fig.~\ref{fig:msplinecurve}.

\subsection{O-splines}
O-splines $O_{i,k}(x|\mathbf{t})$ are introduced to represent square-normalized functions and can be derived by symmetrically orthogonalizing the B-spline basis using L\"owdin's symmetric orthogonalization\cite{Lowding1965AdvPhys}. Let $O_k(x|\mathbf{t}, \bm{\alpha}) = \sum_{i=1}^N  \beta_i O_{i,k}(x|\mathbf{t})$, the square-normalization becomes 
\begin{align}\label{eq:squarenormalized_condition_simplified}
    \int O_k(x|\mathbf{t}, \bm{\alpha})^2 = \int \sum_{i=0}^N \beta_i^2 O_{i,k}^2(x|\mathbf{t}) = 1. 
\end{align}
Due to the orthogonality, one can make each $O_{i,k}(x)$ square normalized, so $\int O_{i,k}^2(x|\mathbf{t}) = 1$, and require the weights $\sum_{i=1}^N\beta_i^2 = 1$. The normalization and orthogonalization are performed once before training.

We sample $O_k(x|\mathbf{t}, \bm{\alpha})^2$ with rejection sampling, which requires the upper bound of$O_k(x|\mathbf{t}, \bm{\alpha})^2$. 
\begin{theorem}
The value of $O_k(x|\mathbf{t}, \bm{\alpha})^2$ is upper bounded by
\begin{align}
     O_k(x|\mathbf{t}, \bm{\alpha})^2 \leq \max{ (\bm{\beta}^T \bm{\Omega})^2},
\end{align}
where $\bm{\Omega}$ denotes the basis change matrix from the B-spline to the O-spline basis. 
\label{theorem:max_value_o_splines}
\end{theorem}

\begin{proof}
 L\"owdin's symmetric orthogonalization defines the relationship 
 $O_{i,k}(x|\mathbf{t}) =\sum_{j} \Omega_{ij}B_{j,k}(x|\mathbf{t}) $, thus
\eqsplits{
    & O_k^2(x|\mathbf{t}, \bm{\alpha})
    =\sum_{i=1}^N \beta_i^2 O_{i,k}^2(x|\mathbf{t})  = 
    \sum_{i=1}^N \beta_i^2 \left(\sum_{j} \Omega_{ij}B_{j,k}(x|\mathbf{t}) \right)^2 \\
    & \leq \sum_{i=1}^N \beta_i^2 [\max_j\Omega_{ij}]^2 \left(\sum_{j=1}^N B_{j,k}(x|\mathbf{t})\right)^2 
    = \sum_{i=1}^N \beta_i^2 [\max_j\Omega_{ij}]^2  \\
    & \leq \max_{i,j}\left(\beta_i \Omega_{ij}\right)^2 
    =  \max{ (\bm{\beta}^T \bm{\Omega})^2}
}
where we used $\sum_{i=1}^N B_{i,k}(x|\mathbf{t}) = 1$.
\end{proof}
A set of O-splines derived from B-splines in Fig.~\ref{fig:bsplines} are shown in Fig.~\ref{fig:obsplines}.

\subsection{I-splines\label{sec:ispline}}
I-spline curves are derived by integrating M-spline curves
\begin{align}
    I_k(x|\mathbf{t}, {\bm{\alpha}}) = \int_0^x M_k(y|\mathbf{t}, {\bm{\alpha}}) \dy
\end{align}
Since M-splines are positive, the integral is monotonically increasing.

Moreover, I-splines are smoother than the M-splines: if the M-spline curve is $k$-differentiable, the corresponding I-spline curve is $(k+1)$-differentiable.
The inverse of $I_k(x|\mathbf{t}, {\bm{\alpha}})$ can be derived by binary search. An example of the I-spine curve is presented in Fig.~\ref{fig:isplinecurve}.


\bibliography{references}

\end{document}